\documentclass[10pt,journal,compsoc]{IEEEtran}
\ifCLASSOPTIONcompsoc
  \usepackage[nocompress]{cite}
\else
  \usepackage{cite}
\fi
\ifCLASSINFOpdf
\else
\fi

\usepackage{subcaption}

\usepackage{graphicx}

\usepackage{tikz}
\usepackage{comment}
\usepackage{amsmath,amssymb} 
\usepackage{multirow}
\usepackage{color}
\usepackage{bm}
\usepackage{booktabs}
\usepackage{bbding}
\newcommand{\mypm}[1]{\color{gray}{\tiny{#1}}}
\usepackage{hyperref}
\usepackage{algorithm}
\usepackage{algorithmicx}
\usepackage{algpseudocode}
\usepackage{cleveref}
\floatname{algorithm}{Algorithm}

\crefname{Algorithm}{Algorithm}{Algorithm}

\begin{document}
%
\title{Towards End-to-end Semi-supervised Learning for One-stage Object Detection}

\author{Gen~Luo, 
	Yiyi~Zhou,~\IEEEmembership{Member,~IEEE,}
	Lei Jin,
	Xiaoshuai Sun,~\IEEEmembership{Member,~IEEE,} \\
	Rongrong Ji,~\IEEEmembership{Senior~Member,~IEEE }
	
	\IEEEcompsocitemizethanks{\IEEEcompsocthanksitem G. Luo,  Y. Zhou,  L. Jin, X. Sun and R. Ji are with the Media Analytics and Computing Lab, Department of Artificial Intelligence, Xiamen University, China (e-mail: luogen@stu.xmu.edu.cn,   zhouyiyi@xmu.edu.cn, kingts@stu.xmu.edu.cn, xssun@xmu.edu.cn, rrji@xmu.edu.cn).}}

%
%

\markboth{Journal of \LaTeX\ Class Files,~Vol.~14, No.~8, August~2015}%
{Shell \MakeLowercase{\textit{et al.}}: Bare Demo of IEEEtran.cls for Computer Society Journals}
%



\IEEEtitleabstractindextext{%
\begin{abstract}
Semi-supervised object detection (SSOD) is a research hot spot in computer vision, which can greatly reduce the requirement for expensive bounding-box annotations.  Despite great success,  existing progress mainly focuses on two-stage  detection networks  like FasterRCNN, while the research on one-stage detectors is often ignored. 
In this paper, we focus on the semi-supervised learning for the advanced and popular one-stage detection network YOLOv5. Compared with Faster-RCNN,  the implementation of YOLOv5 is much more complex, and the various training techniques used in YOLOv5 can also reduce  the benefit of   SSOD.  In addition to this challenge, we also reveal two key  issues in one-stage SSOD, which are \textit{low-quality pseudo-labeling} and \textit{multi-task optimization conflict}, respectively. To address these issues, we propose a novel teacher-student learning recipe called OneTeacher with two innovative designs, namely \textit{Multi-view Pseudo-label Refinement} (MPR) and \textit{Decoupled Semi-supervised Optimization} (DSO). In particular, MPR improves the quality of pseudo-labels via augmented-view refinement and global-view filtering, and DSO handles the joint optimization conflicts via structure tweaks and task-specific pseudo-labeling. In addition, we also carefully revise the implementation of YOLOv5 to maximize the benefits of SSOD, which is also shared with the existing SSOD methods for fair comparison.  To validate OneTeacher, we conduct extensive experiments on COCO and Pascal VOC. The extensive experiments show that OneTeacher   can not only achieve  superior performance than the compared methods, \textit{e.g.,}   15.0\% relative AP gains over  \textit{Unbiased Teacher},  but also  well handle the key  issues  in one-stage SSOD. Code is available at: \url{https://github.com/luogen1996/OneTeacher}.
\end{abstract}

}

\maketitle

\IEEEdisplaynontitleabstractindextext

%
\IEEEpeerreviewmaketitle

\IEEEraisesectionheading{\section{Introduction}\label{sec:introduction}}

%
%
%
%
\IEEEPARstart{R}{ecent} year has witnessed the rapid development of object detection 
aided by a bunch of benchmark datasets~\cite{voc,coco,object365,imagenet} and methods~\cite{rcnn,fastrcnn,fasterrcnn,yolov3,yolox}. Despite       great  success,  the application of object detection has been long plagued  instance-level annotations.
To this end,  numerous efforts~\cite{stac-ssod-ts,unbiasedteacher-ssod-ts,tang2021humbleteacher-ssod-ts,softteacher-ssod-ts} have been devoted to     semi-supervised object detection  (SSOD).

Inspired by the  advances in image classification~\cite{meanteacher-ssl-ts,mixmatch-ssl,fixmatch-ssl}, recent endeavors  also   resort to teacher-student learning for SSOD~\cite{stac-ssod-ts,unbiasedteacher-ssod-ts,instant-ssod-ts,tang2021humbleteacher-ssod-ts,softteacher-ssod-ts}. The main principle of this paradigm is to use a teacher network to generate pseudo labels for the optimization of the student one, where the strong and week data augmentations are separately  applied to enforce the consistency between two networks~\cite{cutout,autoaugment,dataaug_forconsistency-ssl}. This   paradigm can well exploit large amounts of unlabeled data  based on  limited label  information, showing great advantages over  previous semi-supervised approaches~\cite{meanteacher-ssl-ts,fixmatch-ssl,unbiasedteacher-ssod-ts,stac-ssod-ts}.  Therefore, a flurry of  methods~\cite{stac-ssod-ts,unbiasedteacher-ssod-ts,instant-ssod-ts,tang2021humbleteacher-ssod-ts,softteacher-ssod-ts} applying 
teacher-student learning  have been recently proposed to SSOD and achieved   notable success. However, these methods  mainly focus on  two-stage  detection   networks like  FasterRCNN~\cite{fasterrcnn}, while the  exploration on  the  widely-used one-stage models like YOLO series~\cite{yolov3,yolov4,yolov5} has  yet to materialize.

%
\begin{figure*}[t]
	\centering
	\includegraphics[width=0.95\textwidth]{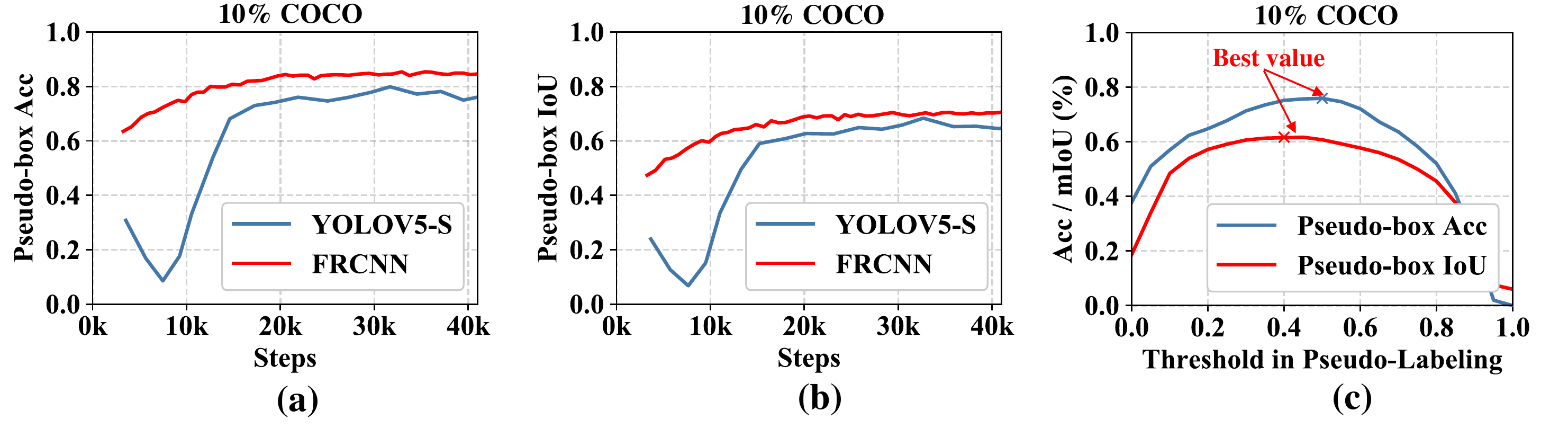}
	\vspace{-1em}
	\caption{ \textbf{  Qualities of pseudo-labels  of YOLOv5~\cite{yolov5} and FasterRCNN~\cite{fasterrcnn}  in  Unbiased Teacher~\cite{unbiasedteacher-ssod-ts}.}   
		The subplots of (a) and (b) show that the average pseudo-label quality of YOLOv5 is much lower than that of FasterRCNN, especially in the initial training phase. Subplot (c) illustrates the different pseudo-label requirements for  box regression and classification.
	}
	\label{fig1}
	\vspace{-1em}
\end{figure*}

Due to the huge gap between  one-stage and two-stage detection networks,   it is   often  a  sub-optimal solution to directly apply existing teacher-student  approaches to one-stage SSOD.  
On one hand, one-stage networks typically adopt a dense prediction paradigm~\cite{yolo,yolo9000,yolov3,yolov5},  which is prone to producing more  noisy pseudo-boxes.  
Concretely, two-stage models like  FasterRCNN  can use multi-stage filterings to ensure the quality of  the predicted bounding boxes, \emph{e.g.}, the proposal selection by RPN  and the further refinement by  ROI head~\cite{fasterrcnn}. However,   one-stage  models    directly enumerate over 100$k$ locations for  prediction.  In this case,  low-quality and noisy  pseudo-boxes  will account for the vast majority when the teacher is not sufficiently trained, as shown in Fig.~\ref{fig1} (a)-(b). 
Therefore, how to select and produce high-quality pseudo-labels   is more critical  in one-stage detectors  than  the two-stage ones.

On the other hand, the joint multi-task optimization of one-stage models also degrades the efficiency of teacher-student learning.  Specifically, two-stage detectors complete the proposal of candidate bounding boxes and the classification of object categories through RPN and RoI head, respectively.  In this case,   different  types of  pseudo-labels   can be   collected for the optimization of each task without conflicts. In contrast, one-stage networks like YOLO series~\cite{yolov3,yolov4,yolo,yolov5,yolo9000} unify these tasks in one  prediction layer, but  the pseudo labels for different objectives are selected by the same criteria in existing SSOD schemes.  Considering that these two tasks usually have different pseudo-label requirements   and noise tolerances, 
 this setting will     exacerbate the    multi-task  optimization conflict  in one-stage detection~\cite{yolox,wu2020rethinking,jiang2018acquisition}, which ultimately  reduces the efficiency of teacher-student learning.

In this paper, we present a  novel teacher-student  learning  approach   for one-stage  SSOD,  termed \textit{OneTeacher}, with two  innovative designs, namely \emph{Multi-view Pesudo-label Refinement} (MPR) and \textit{Decoupled Semi-supervised Optimization} (DSO). 
Specifically, MPR is used to improve the quality of  pseudo-boxes  via  multi-view refinements. It  first employs   augmented-view comparisons to enhance the robustness of pseudo-labels from  the  instance level, and then filters the low-quality ones  according to   image-level  predictions.
In this case, MPR    can help  OneTeacher obtain more reliable pseudo-labels  during training.   Meanwhile,   DSO  is deployed  to address the multi-task optimization conflicts~\cite{yolox}.   It  disentangles  the  joint multi-task optimization with a simple structure modification,  and  also perform task-specific pseudo-labeling to maximize the effect of SSOD.  With these innovative designs, {OneTeacher}  can maximize the benefit of teacher-student learning for one-stage SSOD.

To validate {OneTeacher}, we use YOLOv5~\cite{yolov5} as our base model. As one of the most advanced detection networks, YOLOv5  deploys a series of training techniques in its implementation, such as {EMA}~\cite{meanteacher-ssl-ts}, {cosine learning rate decay} and  {various data augmentations}~\cite{yolov4,yolov3,cutout}. These techniques make  its  implementation  much more complex than  that of FasterRCNN in SSOD~\cite{unbiasedteacher-ssod-ts,stac-ssod-ts,softteacher-ssod-ts,mi2022active}, and some of them   have been used in existing SSOD methods to improve performance~\cite{meanteacher-ssl-ts,cutout} , \emph{e.g.}, EMA and strong data augmentations, which also greatly reduces the benefits of teacher-student learning. 
In this case, we   carefully  revise  the conventional teacher-student  learning settings to make {OneTeacher} applicable to  YOLOv5. For example, we reorganize the data augmentation scheme to accommodate both YOLOv5 and teacher-student learning. Meanwhile, some  hyper-parameters   are also adjusted  according to  the semi-supervised training  statuses  of one-stage models.  
These  modifications are also shared with other SSOD  methods~\cite{stac-ssod-ts,unbiasedteacher-ssod-ts} for fair comparison. 

We  validate our {OneTeacher}  on  two benchmark datasets of object detection, \emph{i.e.,} COCO~\cite{coco} and Pascal VOC~\cite{voc}.  The experimental results show that {OneTeacher} achieves   great improvements over  the supervised baseline, which can be up to +33.5\% relative AP gain on 10\% COCO data. This result is even more significant than that of two-stage SSOD~\cite{unbiasedteacher-ssod-ts,stac-ssod-ts,instant-ssod-ts,softteacher-ssod-ts}. Meanwhile, OneTeacher also achieves better performance than   the state-of-the-art SSOD methods on all experimental settings, \emph{e.g.,} Unbiased Teacher~\cite{unbiasedteacher-ssod-ts}. In addition, extensive quantitative and qualitative analyses also prove that the  issues of pseudo-label quality and optimization conflict are  well handled  by {OneTeacher}.  These results greatly validate the effectiveness of {OneTeacher} towards one-stage SSOD.


In summary, our contribution is three-fold:
\begin{itemize}
	
		\item We identify two key challenges in one-stage semi-supervised object detection, \emph{i.e.}, low-quality pseudo-labels and multi-task optimization conflict, and also realize the first attempt  of SSOD on the advanced one-stage model YOLOv5, of which implementation is much more complex than the two-stage ones.

	\item To address these challenges, we propose a novel  one-stage semi-supervised learning scheme  called OneTeacher with two innovative designs, namely Multi-view Pseudo-label Refinement and Decoupled Semi-supervised Optimization.

	\item  Under all experimental settings, the proposed OneTeacher  obtains  distinct performance gains over both supervised and semi-supervised   methods  on  COCO and Pascal VOC.
\end{itemize}

\begin{figure*}[t] 
	\centering
	\includegraphics[width=0.95\textwidth]{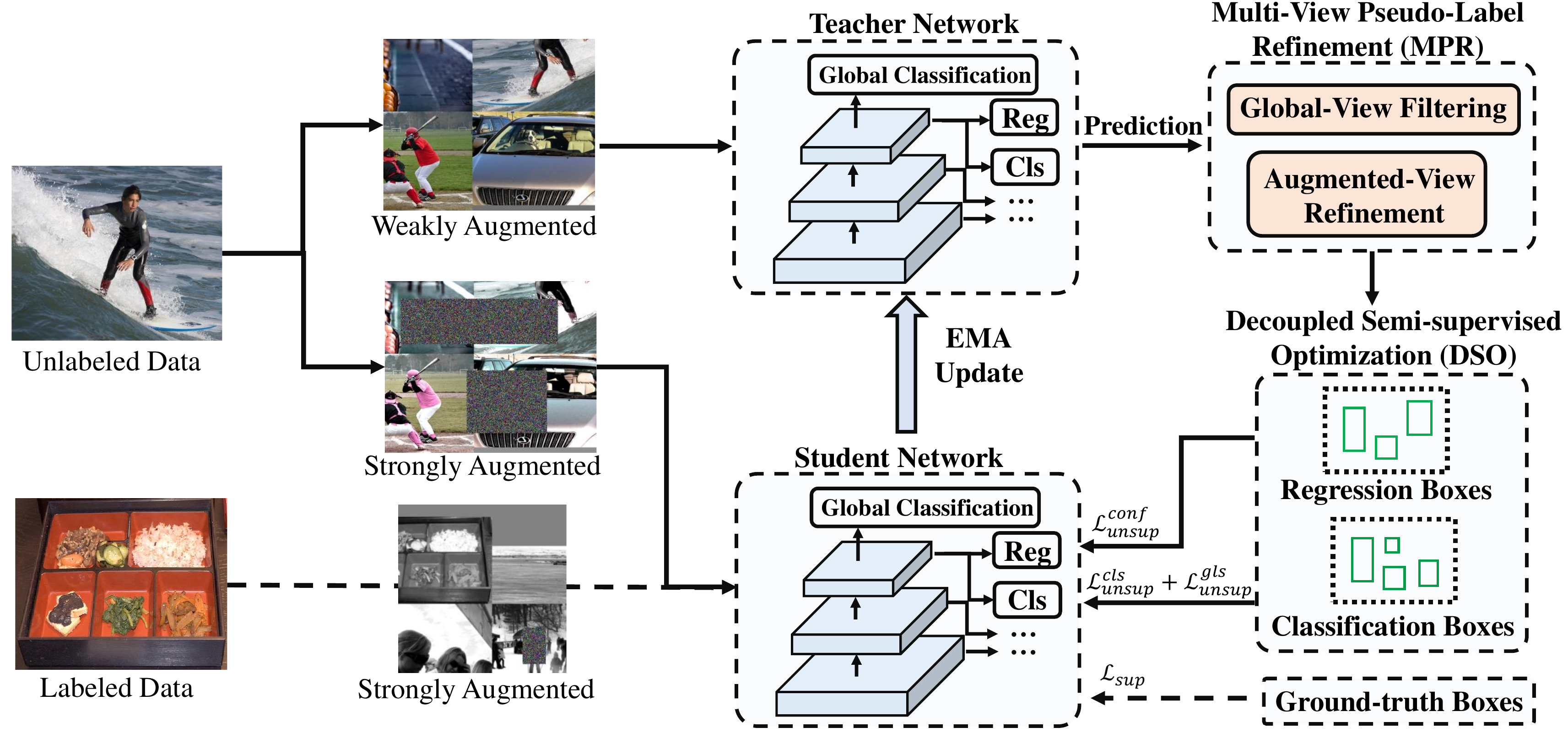}
	\vspace{-1em}
	\caption{\textbf{The  framework of   OneTeacher.} The teacher network   produces high-quality pseudo-labels for  the student  by the proposed    \emph{Multi-view Pseudo-label Refinement (MPR).}  Its parameters are	updated from  the student  ones  via EMA~\cite{meanteacher-ssl-ts}.   OneTeacher also deploys   a novel \emph{Decoupled-Semi-supervised Optimization} (DSO) scheme  to deal with the multi-task optimization conflict~\cite{yolox,wu2020rethinking}. The base model used is YOLOv5~\cite{yolov5}.  }
	\label{framework} 
\end{figure*}

\section{Related Work}
\subsection{Object Detection}
 Object detection is a fundamental task in computer vision~\cite{yolo,yolo9000,yolov3,ssd,rcnn,fpn,fastrcnn,fasterrcnn,maskrcnn,cascadercnn}. In the era of deep learning, its modeling strategies can be roughly categorized into two main groups, \emph{i.e.}, the  two-stage and one-stage approaches~\cite{yolo,yolo9000,yolov3,ssd,rcnn,fastrcnn,fasterrcnn,maskrcnn,fpn,yolox}. Specifically, two-stage approaches~\cite{rcnn,fastrcnn,fasterrcnn,maskrcnn,cascadercnn} first generate a manageable number of candidate object regions, and then the features of these regions are pooled  to predict the corresponding box coordinates and categories.  In practice, two-stage methods such as FasterRCNN~\cite{fasterrcnn} and Cascade RCNN~\cite{cascadercnn} often exhibit better robustness and performance than the single-stage ones, but their inference   is less efficient.  Compared to two-stage approaches, one-stage approaches~\cite{ssd,yolo,yolo9000,yolov3,ssd,yolox,yolov5,centernet} directly predict the coordinates and categories of objects based on the  convolutional feature maps.  Aided by the simple and flexible structure,  one-stage approaches can easily achieve a good trade-off  between performance and inference speed~\cite{yolov3,yolo,yolo9000,ssd}.  To this end, the study of one-stage  detection  has  gained  increasing attention, which leads to the birth of a flurry of innovative detection networks~\cite{ssd,yolo,yolo9000,yolov3,ssd,yolox,yolov5,centernet}, such as CenterNet~\cite{centernet} and YOLO-series~\cite{yolo,yolo9000,yolov3,yolox,yolov5}.  In this paper, we focus on the semi-supervised learning  for  one of the most advanced one-stage detectors called  YOLOv5~\cite{yolov5}. Compared   with previous  YOLO models~\cite{yolov3,yolov4,yolo9000,yolo}, YOLOv5  applies a  stronger visual backbone and  also involves  more training techniques, such as EMA and data augmentations, which makes its implementation of SSOD more difficult than the two-stage detection networks like Faster-RCNN. 

\subsection{Semi-supervised Learning}
Semi-supervised learning (SSL) aims to train a model with   limited label information and massive amounts of unlabeled data.   In computer vision, the research of SSL  on image classification  garners an influx of interest~\cite{consistency-based-ssod,consistency-ssl,mixmatch-ssl,remixmatch-ssl,meanteacher-ssl-ts,fixmatch-ssl,learning-with-pseudo-ensembles-ssl}.   In early SSL methods~\cite{mixmatch-ssl,remixmatch-ssl,meanteacher-ssl-ts,fixmatch-ssl}, one popular solution is  to apply  consistency regularization, which encourages the model to make consistent predictions  with different perturbations, \emph{e.g.,} data augmentations~\cite{meanteacher-ssl-ts}, stochastic regularization~\cite{temporal-ssl,sajjadi2016regularization} and adversarial perturbations~\cite{miyato2018virtual}.  Another direction of SSL is  pseudo-label based learning~\cite{arazo2020pseudo,pham2019semi}, which  uses the predictions of the model as  hard labels for semi-supervised  training.  Recently, researchers  resort to teacher-student learning and combine the merits of these two methodologies for better SSL~\cite{meanteacher-ssl-ts,mixmatch-ssl,remixmatch-ssl,fixmatch-ssl}. In particular, Mean Teacher~\cite{meanteacher-ssl-ts}  applies data augmentation to the student and calculates  its  consistency loss with the teacher. To improve the training  stability, it  also employs   \textit{exponential moving average} (EMA) to    update  the teacher's parameters  from the student ones, which can  prevent the model from confirmation bias~\cite{unbiasedteacher-ssod-ts}.   MixMatch~\cite{mixmatch-ssl} applies $k$  types of  different stochastic data augmentations to unlabeled images, and averages their predictions as the final pseudo-labels.  Based on MixMatch~\cite{mixmatch-ssl}, ReMixMatch~\cite{remixmatch-ssl} proposes distribution alignments to align the prediction distributions of unlabeled and labeled data. It also proposes strong-weak augmentation strategies to ensure the consistency between the teacher and  the  student. 
More recently, some  methods also aim to filter the noisy pseudo-labels by the prediction probability~\cite{fixmatch-ssl}, and  address the class-imbalanced problem~\cite{guo2022class}.

%
%
%

\begin{figure*}[t]
	\centering
	\includegraphics[width=1.0\textwidth]{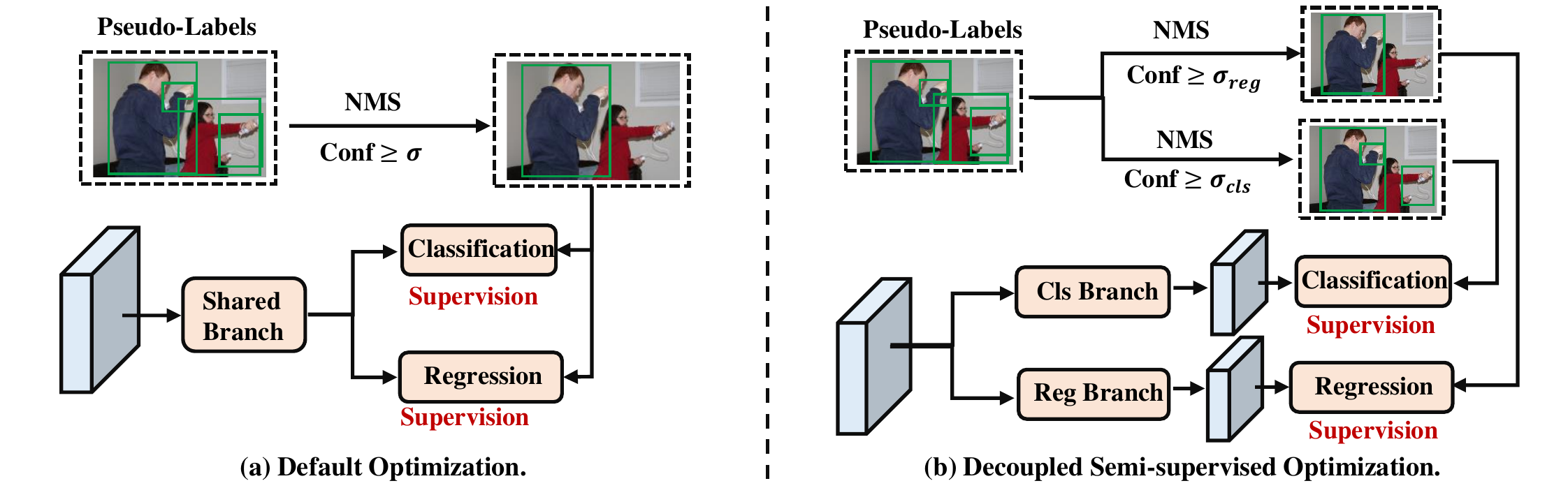} 
	\vspace{-2em}
	\caption{\textbf{Comparison of  the default  optimization  of YOLOv5 and the proposed  Decoupled Semi-supervised Optimization (DSO).} The default optimization scheme  will exacerbates multi-task optimization conflicts~\cite{yolox,wu2020rethinking} during teacher-student learning. The proposed DSO addresses this issue via a  structure tweak and task-specific pseudo-labeling. }
	\label{decouple} 
\end{figure*} 

\subsection{Semi-supervised Object Detection}

Due to the expensive instance-level annotations, semi-supervised object detection (SSOD) has long been a research spot in computer vision.  Early SSOD approaches  such as CSD~\cite{consistency-based-ssod} and ISD~\cite{jeong2021interpolation-ssod}   introduce the consistency-based  learning schemes to object  detection networks. Recently, the teacher-student framework~\cite{meanteacher-ssl-ts} becomes  popular in SSOD.  The teacher network  is used to produce pseudo-boxes, and the student network is trained with both labeled and pseudo-boxes.  In particular, STAC~\cite{stac-ssod-ts} is a  representative teacher-student based SSOD approach,  which divides the SSL into two steps, \emph{i.e.},  the pseudo-labeling process for unlabeled data and the re-training step based on the pseudo-labels.   Despite the effectiveness, STAC still suffers critical  imbalance issue and overfitting. To address these issues, Unbiased Teacher~\cite{unbiasedteacher-ssod-ts} proposes an end-to-end SSOD method, which introduces EMA~\cite{meanteacher-ssl-ts} to update the parameters of the teacher network from the student  one, and  applies Focal loss~\cite{lin2017focalloss} to address  the issue of error accumulation.   In addition to Unbiased Teacher, a set of SSOD approaches~\cite{instant-ssod-ts,tang2021humbleteacher-ssod-ts,softteacher-ssod-ts,mi2022active} are proposed recently. Among
 them, most work focuses on the strategies  of improving the  pseudo-label  quality, \emph{e.g.,} the co-rectify strategy~\cite{instant-ssod-ts}, the detection-specific data  ensemble  strategy~\cite{tang2021humbleteacher-ssod-ts} and the  box jittering approach~\cite{softteacher-ssod-ts}.  Different from these work, Active Teacher~\cite{mi2022active}  aims to select  the optimal labeled examples for  SSOD via three   metrics, \emph{i.e.,}  \textit{difficulty}, \textit{information} and \textit{diversity}.    Overall,  most existing  teacher-student based  approaches are designed for  the  two-stage detector FasterRCNN.


 Very recently, some  advances~\cite{chen2022dense,liu2022unbiased} also  explore SSOD  for  one-stage detection networks like  FCOS~\cite{tian2019fcos}.  In particular, Liu \textit{et.al}~\cite{liu2022unbiased} propose the modeling of localization uncertainty to select reliable pseudo-labels.   Chen \textit{et.al}~\cite{chen2022dense}  improve the pseudo-label alignments and qualities for one-stage SSOD via three novel designs, \emph{i.e.,} an adaptive filtering strategy,  an aggregated teacher and an uncertainty-consistency-regularization strategy.
In this paper, we  are committed  to exploring the     teacher-student learning paradigm  for  advanced one-stage detection networks, \emph{i.e.,}  YOLOv5 \cite{yolov5}.  Compared to these works, this paper elaborates the great gap between one-stage and two-stage SSODs in more detail and depth, especially for the popular YOLO series. Meanwhile, we also investigate  how to make  advanced training techniques used in one-stage detectors to be compatible with teacher-student learning.

\section{OneTeacher}
\subsection{ Overview}  The framework of the proposed OneTeacher is depicted in Fig.~\ref{framework}, which consists  of two  detection  networks with the same configurations, namely \emph{teacher} and \emph{student}. The teacher  network   is in charged of generating  pseudo-labels, based on which the student  one is trained together with the  ground-truth   labels.  To this end, the optimization of the student network is defined by 
\begin{equation}
\mathcal{L} = \mathcal{L}_{sup} + \lambda \cdot \mathcal{L}_{unsup},
\label{eq:semi-loss}
\end{equation}
where $\lambda$ is the hyper-parameter to adjust the  contribution  of unsupervised loss. 
During  training, the parameters of  the  teacher network  $\theta_t$ are updated  from the student  ones  $\theta_s$ via   \textit{exponential moving average} (EMA)~\cite{meanteacher-ssl-ts}, which can be formulated as
\begin{equation}
\label{equ:train-equ}
\begin{aligned}
&\theta_s \leftarrow \theta_s + \gamma \frac{\partial (\mathcal{L}_{sup} + {\lambda}_u \mathcal{L}_{unsup}) }{\partial\theta_s}, \\
&\theta_{t} \leftarrow \alpha \theta_{t} + (1-\alpha) \theta_{s}.
\end{aligned}
\end{equation}
Here,  $\gamma$  denotes the learning rate and  $\alpha$ is the   EMA coefficient.  The use of EMA is  to allow the teacher network to generate stable pseudo-labels during training,  thereby alleviating the effects of  pseudo-label bias~\cite{unbiasedteacher-ssod-ts}.
In practice, the teacher network can  be regarded as an ensemble of the student networks  in different training statuses, which is  also  used as the target model after training.

\begin{figure*}[t]
	\centering
	\includegraphics[width=1.0\textwidth]{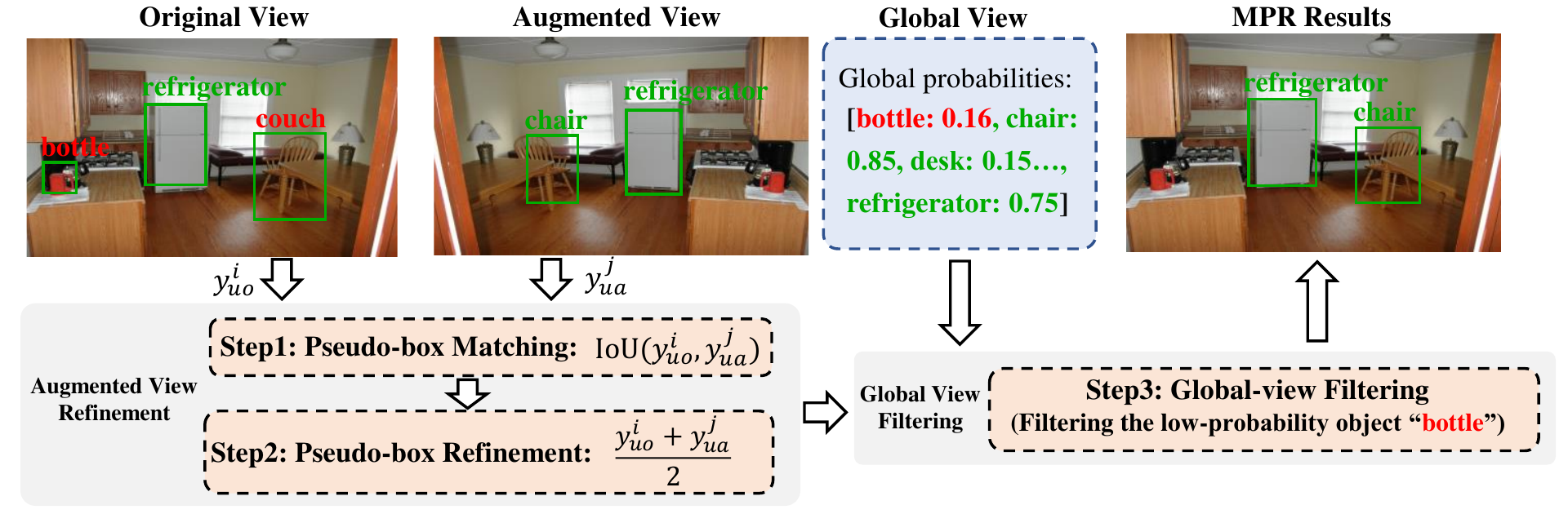}
	\vspace{-2em}
	\caption{\textbf{Illustration of Multi-view Pseudo-Label Refinement (MPR).} In MPR, the pseudo-labels are first refined by fusing the   matched predictions from the augmented view.  The  global-view filtering is then performed to  remove the  pseudo-labels  of the categories   that lower than the global prediction threshold.   }
	\vspace{-1em}
\end{figure*}
\subsection{ Semi-supervised Learning for One-stage Object Detection} 

One-stage detection networks often adopt an end-to-end prediction paradigm, which is obviously different from the two-stage ones.  Specifically, given  a labeled  image $I_l \in \mathbb{R}^{H \times W \times 3 }$, the one-stage detection network  like the YOLO series~\cite{yolov3,yolov4,yolov5} will   output  an tensor  $t_l \in \mathbb{R}^{h\times w\times (c+5)}$ for  the joint multi-task prediction, where $c$ denotes the number of object categories and 5 refers to  the confidence score and the coordinates of  the bounding box.   To this end, the supervised  loss  $\mathcal{L}_{sup}$ can be   defined by
\begin{equation}
\label{equ:full_sup}
\begin{aligned}
\mathcal{L}_{sup}(t_l,Y_l)&=\mathcal{L}_{iou}(t_l, Y_l^{coor}) + \mathcal{L}_{conf}(t_l, Y_l^{conf}) \\&+ \mathcal{L}_{cls}(t_l,Y_l^{cls}),
\end{aligned}
\end{equation}
where $Y_l=\{Y_l^{cls}, Y_l^{conf}, Y_l^{coor}\}$ are the   labels of categories, confidence and coordinates. $\mathcal{L}_{cls}$ and $\mathcal{L}_{conf}$   are  the binary cross entropy losses for category classification  and confidence regression,  and  $\mathcal{L}_{iou}$ denotes the IoU loss for bounding box regression. 

In terms of  unsupervised loss, a straightforward solution is to select the predicted bounding boxes above a fixed threshold as pseudo-labels, which  can be formulated as
\begin{equation}
\label{equ:full_unsup}
\begin{aligned}
\mathcal{L}_{unsup}(t_u,Y_u)&=\mathcal{L}_{iou}(t_u, Y_u^{coor}) + \mathcal{L}_{conf}(t_u, Y_{u}^{conf}) \\&+ \mathcal{L}_{cls}(t_u, Y_{u}^{cls}).
\end{aligned}
\end{equation}
where $Y_u=\{Y_u^{cls}, Y_u^{conf}, Y_u^{coor}\}$ are the pseudo-label  set of  object categories, confidence and coordinates\footnote{In some methods,  bounding box regression is ignored in unsupervised loss.}.

A potential problem   is the optimization conflict between regression and classification in one-stage detection, as revealed in \cite{yolox,wu2020rethinking}. This issue becomes more severe in semi-supervised learning. To explain, pseudo-labels are of much lower quality compared to the  ground-truth ones. Meanwhile, classification and regression have different requirements about pseudo-labels, which  are selected by the same threshold in existing methods~\cite{unbiasedteacher-ssod-ts,instant-ssod-ts,stac-ssod-ts}.

\subsubsection{Decoupled semi-supervised optimization.} \label{DSO_sec}  To alleviate    multi-task optimization conflict, we propose   a novel  \textit{Decoupled Semi-supervised Optimization} (DSO)  scheme for {OneTeacher}, which decouples  the  joint optimization via a simple  branching structure  and a task-specific pseudo-labeling  strategy.

As shown in Fig.~\ref{decouple},  for each  image,   we decompose the prediction branch into two separate ones, and then obtain the prediction tensors  $t_l^{cls} \in \mathbb{R}^{h\times w\times c}$ and $t_l^{reg} \in \mathbb{R}^{h\times w\times 5}$ for  classification and regression, respectively.   Thus, the supervised loss  defined  in  E.q~\ref{equ:full_sup} can be reformulated as
\begin{equation}
\label{equ:supervised-equ-detail}
\small
\begin{aligned}
\mathcal{L}_{sup}(t_l^{cls},t_l^{reg},Y_l)&=\mathcal{L}_{iou}(t_l^{reg}, Y_l^{coor}) + \mathcal{L}_{conf}(t_l^{reg}, Y_l^{conf}) \\&+ \mathcal{L}_{cls}(t_l^{cls},Y_l^{cls}).
\end{aligned}
\end{equation}
This   modification can greatly avoid conflicts across tasks.

Afterwards, we perform task-specific pseudo-labeling for  unsupervised loss.   Specifically, given the predictions of an unlabeled image by the teacher, we use the multiplication of the confidence  probability  and the  max classification score as the  indicator. Based on this indicator,  we set   two different thresholds $\sigma_{reg}$ and $\sigma_{cls}$    to select pseudo-labels  for regression and classification,  denoted as $Y_{ur}$  and $Y_{uc}$,  respectively.     Then,   these  two  tasks are trained with the corresponding pseudo-labels , and the  the unsupervised  loss of {OneTeacher}  can be written as
\begin{equation}
\label{equ:unsupervised-equ-detail}
\begin{aligned}
\mathcal{L}_{unsup}(t^{cls}_u,t^{reg}_u,Y_{ur},Y_{uc})
&=\mathcal{L}_{conf}(t_u^{reg}, Y_{ur}^{conf}) \\&+ \mathcal{L}_{cls}(t_u^{cls}, Y_{uc}^{cls}).
\end{aligned}
\end{equation}
Here, we follow the setting of \cite{unbiasedteacher-ssod-ts} to discard the unsupervised optimization of bounding box regression.    
This  task-specific pseudo-labeling  strategy can  flexibly   adjust the noisy degree of different tasks, thereby improving the  efficiency of teacher-student learning. 

During deployment, we also add a multi-label classification task to the model, which is used in the pseudo-label refinement  discussed in Sec.~\ref{MPR}. In this case, the final objective function of \textit{OneTeacher} is defined as 
\begin{equation}
\label{equ:loss-equ-detail-final}
\begin{aligned}
\mathcal{L}&=\mathcal{L}_{sup}(t_l^{cls},t_l^{reg}, t_l^{gls}, Y_l) \\&+\lambda \cdot \mathcal{L}_{unsup}(t_u^{cls}, t_u^{reg}, t_u^{gls}, Y_{ur},  Y_{uc}),
\end{aligned}
\end{equation}
where $t^{gls}$ is the  prediction tensor  of the multi-label classification branch. $Y_{ur}$ and $Y_{uc}$ are the pseudo-labels selected  by different thresholds.

\begin{algorithm}[t]
	\caption{Pseudo Code of Augmented-view Refinement}  
	\label[Algorithm]{alg:pseudocode_of_aug_view}  
	\begin{algorithmic}[1]  
		\Require  
		Pseudo-labels of the original view $Y_{uo}$ and the augmented view $Y_{ua}$.
		\Ensure  
		Refined Pseudo-labels $Y_u$.
		\ForAll{$y_{uo}^i \in Y_u^{ori}$}
		\ForAll{$y_{ua}^j \in Y_u^{aug}$}
		\State Calculating IoU values   between $y_{uo}^i$ and  $y_{ua}^j$:
		\State  $\text{iou}_{oa}^{j}$=IoU($y_{uo}^i $, $y_{ua}^j$)
		\EndFor
		\State Selecting $y_{ua}^j$ with the largest $\text{iou}_{oa}^{j}$
		\If  {  $\text{iou}_{oa}^{j}$ $>$$\sigma$} 
		\State $y_u= \frac{y_{uo}^i+y_{ua}^j}{2}$
		\Else  
		\State $y_u=y_{uo}^i$ 	 
		\EndIf 
		\EndFor \\
		\Return $\{y_u^{1},...,y_u^{n}\}$
	\end{algorithmic}
\end{algorithm} 

\subsection{Multi-view Pseudo-label Refinement} In two-stage networks like FaterRCNN, the teacher's predictions will be screened by RPN and ROI head successively to obtain the final pseudo-label set~\cite{unbiasedteacher-ssod-ts,instant-ssod-ts,stac-ssod-ts}. This multi-step selection can somewhat ensure the qualities of pseudo labels, which is however not applicable to one-stage models due to different prediction paradigms.
A compromise solution for one-stage models is to directly use their confidence scores to determine pseudo-labels,  which is  insufficient for  evaluating the quality of pseudo-labels. 

\label{MPR}
To address this issue, we propose a novel \textit{ Multi-view Pseudo-label Refinement} (MPR)   scheme, which consists of two main processes, namely  \textit{augmented-view refinement} and  \textit{global-view filtering}. The processing of MPR are shown in Algorithm \ref{alg:pseudocode_of_aug_view}. Specifically, given an unlabeled image $I_u$, we first apply augmented-view refinement to adjust its pseudo-label information.
As shown in Fig.~\ref{MPR}, the teacher network will predict the bounding boxes of $I_u$ and its augmented view of flipping, denoted as $Y_{uo}$ and $Y_{ua}$, respectively.  
Afterwards, we compare each bounding box in $Y_{uo}$ with the one in $Y_{ua}$ by \textit{IoU} value, and select the matched bounding boxes from two views, denoted as $(y_{uo}, y_{ua})$.

  In practice, this process consists of  two steps. Firstly, given a $y_{uo}^i$, we find the  $y_{ua}^j$ that have the max IoU value with $y_{uo}^i$, which can be formulated as:
  \begin{equation} 
  \begin{aligned}
  j=\mathop{\arg \max}_{y_{ua}^j \in Y_{ua}, y_{uo}^i \in Y_{uo}} \text{IoU}(y_{ua}^j,y_{uo}^i).
  \end{aligned}
  \end{equation}
   Then, if the IoU value is large than a threshold $\delta$, they are regarded as a pair $(y_{uo}^i,y_{ua}^j)$. In practice, the threshold is set to 0.45.  Lastly, the final pseudo box $y_u^i$ averages the information of $(y_{uo}^i, y_{ua}^j)$ including confidence scores, classification probabilities and bounding box coordinates, which can be written by
     \begin{equation} 
   \begin{aligned}
  y_u^i=\frac{y_{uo}^i+ y_{ua}^j}{2}.
   \end{aligned}
   \end{equation}
    The intuition behind  this   augmented view refinement is that  the ensemble of predictions can  alleviate   the bias of the teacher  to some extent, thereby making the final pseudo-labels more reliable.

\begin{table}[t]
	\centering
	\caption{\textbf{The  implementation   of YOLOV5  for OneTeacher. The first block shows different training settings. } ``Opt.'' denotes the optimization strategies. ``ImageNet init'' denotes the pre-training on ImageNet.  ``BR Aug.'' and ``BIR Aug.'' denote the box-relevant and box-irrelevant  data  augmentations, respectively.  }
	\setlength 	\tabcolsep{4pt}
	\begin{tabular}{ll|ccc}
		\toprule[1.2pt]
		&{Method}       &FRCNN    & YOLOV5       & OneTeacher   \\ \midrule[0.8pt]
		\multirow{3}{*}{Opt.}	&{EMA~\cite{meanteacher-ssl-ts} }  &   $\times$             & $\checkmark$ & $\checkmark$ \\
		&{ImageNet init} & $\checkmark$      & $\times$     & $\times$     \\
		&{Learning rate decay} & step & cosine       & constant     \\ \midrule[0.8pt]
		\multirow{3}{*}{BR Aug.} & 	Flip-lr   &   $\checkmark$     & $\checkmark$ & $\checkmark$ \\
		&	Mosaic~\cite{yolov4}      & $\times$       & $\checkmark$ & $\checkmark$ \\
		&Scale Jitter     & $\times$       & $\checkmark$ & $\checkmark$ \\ \midrule[0.8pt]
		\multirow{6}{*}{BIR Aug.} &	HSV translating  & $\times$       & $\checkmark$ & $\checkmark$ \\
		&	 	Image translating   & $\times$     & $\checkmark$ & $\checkmark$ \\
		&		 ColorJitter~\cite{yolov3}  & $\times$         & $\times$     & $\checkmark$ \\
		&	 	Grayscale   & $\times$             & $\times$     & $\checkmark$ \\
		&		 GaussianBlur   & $\times$          & $\times$     & $\checkmark$ \\
		&	 	RandomErasing~\cite{zhong2020random}   & $\times$         & $\times$     & $\checkmark$ \\ \bottomrule[1.2pt]
	\end{tabular}
	\label{tab_adap}
\end{table}
Inspired by the recent SSOD methods~\cite{zhang2021semi}, we also introduce an additional multi-label classification to enhance MPR from a global-view filtering. Concretely,  the teacher network will     output the image-level  multi-class probability  distribution.   If the global probability of a  specific  category is lower than a threshold $\sigma_g$, we will filter the pseudo-boxes of this class. Our assumption  is that the category recognition of local pseudo bounding boxes  should be consistent  with the global one, otherwise these pseudo boxes are often inferior in quality.  

 Overall, the proposed MPR can filter a vast amount of low-quality bounding boxes  and also improve the quality of pseudo-labels greatly.
After MPR, we   apply task-specific thresholds to select the final pseudo-label sets for classification and regression, as discussed in Sec~\ref{DSO_sec}.

\subsection{ Implementation on YOLOv5} \label{imp}
To validate the proposed {OneTeacher}, we further apply it to YOLOv5~\cite{yolov5}, one of the most advanced one-stage detection networks. However, it is not feasible to directly apply  existing teacher-student learning based SSOD   methods~\cite{stac-ssod-ts,unbiasedteacher-ssod-ts} to YOLOV5, which is mainly attributed to the  conflicts of training techniques used by YOLOV5 and SSOD. This  problem is usually ignored in existing SSOD approaches since the default implementation of two-stage detection networks like Faster-RCNN is relatively simple~\cite{fasterrcnn}. However, in real-world applications, a set of training techniques are usually used  to improve  model performance and robustness. 

The first problem encountered is the data augmentation setup.  In  YOLOv5~\cite{yolov5} and  other  advanced  one-stage    detectors~\cite{yolov4,yolox}, strong data augmentations like  \textit{color jittering}~\cite{yolov3} have become the \emph{de facto} training setting, which, however,   are also used as a solution to enforce the consistency learning of the student network in SSOD~\cite{unbiasedteacher-ssod-ts,instant-ssod-ts}.   Abandoning these data augmentations will inevitably degrade the performance of YOLOv5, especially considering that YOLOv5 is trained from scratch without ImageNet pre-training.
A direct solution   is to keep  the data augmentation  settings unchanged and deploy  additional   augmentations  for the student network. But the problem that comes with it is that the teacher will inevitably produce lower-quality labels after receiving the images with strong perturbations.

\begin{figure}[t]
	\centering
	\includegraphics[width=0.48\textwidth]{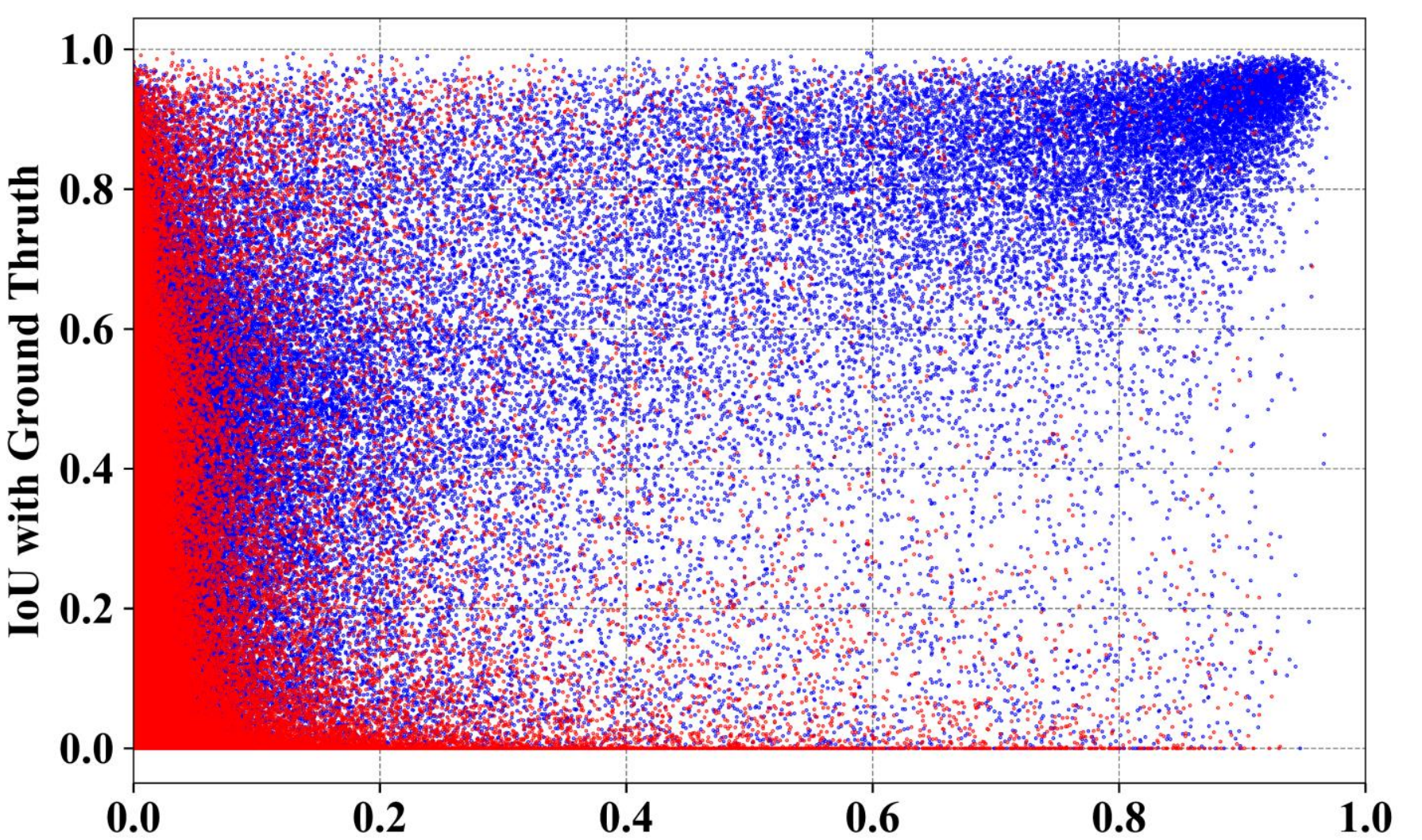} 
	\caption{\textbf{Confidence score distribution of OneTeacher on COCO (10\%). } The blue and red points refer to the correct and wrong predictions, respectively. A threshold of 0.4 can achieve the best trade-off between   quality and quantity. }
	\label{conf_distri_fig}
\end{figure}

\begin{table*}[t]
	\centering
	\caption{\textbf{Comparison of the proposed {OneTeacher} and other SSOD methods on {COCO}.} The ``supervised'' denotes  the fully supervised baseline.  We use {YOLOv5-s} as the  base model and report the mAP scores on COCO \textit{val2017}.}
	\setlength\tabcolsep{10pt}
	\vspace{-2mm}
	\scalebox{0.95}[0.95]{
		\begin{tabular}{lcccccccccc}
			\toprule[1.2pt]
			\multicolumn{11}{c}{COCO}                                      \\ \midrule[0.8pt]
			\multicolumn{1}{c|}{\multirow{2}{*}{Model}} & \multicolumn{2}{c|}{1\%}                              & \multicolumn{2}{c|}{2\%}                               & \multicolumn{2}{c|}{5\%}                               & \multicolumn{2}{c|}{10\%}                              & \multicolumn{2}{c}{20\%}          \\
			\multicolumn{1}{c|}{}                       & mAP            & \multicolumn{1}{c|}{AP50}            & mAP             & \multicolumn{1}{c|}{AP50}            & mAP             & \multicolumn{1}{c|}{AP50}            & mAP             & \multicolumn{1}{c|}{AP50}            & mAP             & AP50            \\ \midrule[0.8pt]
			\multicolumn{1}{l|}{Supervised~\cite{yolov5}}             & 5.2\mypm{+0.0} & \multicolumn{1}{c|}{10.2\mypm{+0.0}} & 9.1\mypm{+0.0}  & \multicolumn{1}{c|}{17.7\mypm{+0.0}} & 15.4\mypm{+0.0} & \multicolumn{1}{c|}{28.5\mypm{+0.0}} & 21.8\mypm{+0.0} & \multicolumn{1}{c|}{37.4\mypm{+0.0}} & 26.7\mypm{+0.0} & 44.2\mypm{+0.0} \\ \midrule[0.8pt]
			\multicolumn{1}{l|}{STAC~\cite{stac-ssod-ts}}                   & 6.4\mypm{+1.2} & \multicolumn{1}{c|}{12.9\mypm{+2.7}} & 10.6\mypm{+1.5} & \multicolumn{1}{c|}{20.4\mypm{+2.7}} & 14.2\mypm{-1.2} & \multicolumn{1}{c|}{27.5\mypm{-1.0}} & 18.3\mypm{-3.5} & \multicolumn{1}{c|}{34.7\mypm{-2.7}} & 21.7\mypm{-5.0} & 40.0\mypm{-4.2} \\
			\multicolumn{1}{l|}{STAC~\cite{stac-ssod-ts}+EMA}               & 8.2\mypm{+3.0} & \multicolumn{1}{c|}{15.2\mypm{+5.0}} & 14.1\mypm{+5.0} & \multicolumn{1}{c|}{25.5\mypm{+7.8}} & 20.7\mypm{+5.3} & \multicolumn{1}{c|}{36.1\mypm{+7.6}} & 25.6\mypm{+3.8} & \multicolumn{1}{c|}{43.0\mypm{+5.6}} & 29.3\mypm{+2.6} & 48.1\mypm{+3.9} \\
			\multicolumn{1}{l|}{Unbiased Teacher~\cite{unbiasedteacher-ssod-ts}}       & 8.3\mypm{+3.1} & \multicolumn{1}{c|}{14.8\mypm{+4.6}} & 14.0\mypm{+4.9} & \multicolumn{1}{c|}{25.5\mypm{+7.8}} & 20.7\mypm{+5.3} & \multicolumn{1}{c|}{35.4\mypm{+6.9}} & 25.3\mypm{+3.5} & \multicolumn{1}{c|}{42.0\mypm{+4.6}} & 30.2\mypm{+3.5} & 48.6\mypm{+4.4} \\ \midrule[0.8pt]
			\multicolumn{1}{l|}{OneTeacher}             & \textbf{8.6}\mypm{+3.4} & \multicolumn{1}{c|}{\textbf{15.4}\mypm{+5.2}} & \textbf{14.7}\mypm{+5.6} & \multicolumn{1}{c|}{\textbf{25.7}\mypm{+8.0}} & \textbf{22.9}\mypm{+7.5} & \multicolumn{1}{c|}{\textbf{36.7}\mypm{+8.2}} & \textbf{29.1}\mypm{+7.3} & \multicolumn{1}{c|}{\textbf{45.3}\mypm{+7.9}} & \textbf{33.2}\mypm{+6.5} & \textbf{50.9}\mypm{+6.7} \\ \bottomrule[1.2pt]
	\end{tabular}}
	\label{sota}
\end{table*}

\begin{table}[t]
	\centering
	\caption{\textbf{Comparison of {OneTeacher} and  the baselines  on Pascal VOC.}  COCO20 denotes the images of the same categories as VOC in COCO \emph{train2017}.}
	\setlength\tabcolsep{5pt}
	\vspace{-2mm}
	{
		\begin{tabular}{lcccc}
			\toprule[1.2pt]
			\multicolumn{5}{c}{VOC}                                                                                                 \\ \midrule[0.8pt]
			\multicolumn{1}{c|}{Methods}          & Labeled & \multicolumn{1}{c|}{Unlabeled}    & mAP             & AP50            \\ \midrule[0.8pt]
			\multicolumn{1}{l|}{Supervised}       & VOC07   & \multicolumn{1}{c|}{None}         & 41.3\mypm{+0.0}            & 66.0\mypm{+0.0}             \\ \midrule[0.8pt]
			\multicolumn{1}{l|}{STAC~\cite{stac-ssod-ts}}             & VOC07   & \multicolumn{1}{c|}{VOC12}        & 38.9\mypm{-2.4} & 66.8\mypm{+1.8} \\
			\multicolumn{1}{l|}{STAC~\cite{stac-ssod-ts}+EMA}         & VOC07   & \multicolumn{1}{c|}{VOC12}        & 44.5\mypm{+3.2} & 70.5\mypm{+4.5} \\
			\multicolumn{1}{l|}{UbTeacher~\cite{unbiasedteacher-ssod-ts}}        & VOC07   & \multicolumn{1}{c|}{VOC12}        & 42.1\mypm{+0.8} & 68.0\mypm{+2.0}   \\
			\multicolumn{1}{l|}{OneTeacher(ours)} & VOC07   & \multicolumn{1}{c|}{VOC12}        & \textbf{45.3}\mypm{+4.0} & \textbf{70.8}\mypm{+4.8} \\ \midrule[0.8pt]
			\multicolumn{1}{l|}{STAC~\cite{stac-ssod-ts}}             & VOC07   & \multicolumn{1}{c|}{VOC12+COCO20} & 38.4\mypm{-2.9} & 66.4\mypm{+0.4} \\
			\multicolumn{1}{l|}{STAC+EMA~\cite{stac-ssod-ts}}         & VOC07   & \multicolumn{1}{c|}{VOC12+COCO20} & 44.7\mypm{+3.4} & 70.8\mypm{+4.8} \\
			\multicolumn{1}{l|}{UbTeacher~\cite{unbiasedteacher-ssod-ts}}        & VOC07   & \multicolumn{1}{c|}{VOC12+COCO20} & 44.2\mypm{+2.9} & 70.2\mypm{+4.2} \\
			\multicolumn{1}{l|}{OneTeacher(ours)} & VOC07   & \multicolumn{1}{c|}{VOC12+COCO20} & \textbf{46.1}\mypm{+4.8} &\textbf{ 71.4}\mypm{+5.4} \\ \bottomrule[1.2pt]
	\end{tabular}}
	\label{voc}
\end{table}

To address this issue, we  categorize data augmentations into two groups, \emph{i.e.}, the \textit{box-relevant } and \textit{box-irrelevant} augmentations, as shown in Tab.~\ref{tab_adap}.  Particularly, box-relevant augmentations, such as \textit{flip} and \textit{mosaic}~\cite{yolov4},  can augment the  bounding box  information, while  having less impact on the image representation. On the contrary, box-irrelevant  methods  will not affect the ground-truth boxes but strongly perturb image content, such as  \textit{color transformation} and \textit{Gaussian blur}.  
Therefore, we  keep box-relevant  methods as the weak data augmentation for the teacher, and use both box-relevant and  box-irrelevant  methods as the strong data augmentation for the student. This strategy can  minimize the  perturbations to the teacher network, while preserving the  original  settings of YOLOv5.

In addition, we  also adjust some common hyper-parameters in  teacher-student learning  for YOLOv5. Specifically, we lower the threshold for pseudo-labeling  to 0.4, which is often set as a high value in two-stage SSOD~\cite{unbiasedteacher-ssod-ts,instant-ssod-ts}, \emph{e.g.}, 0.7. This change is   attributed  to  the noisy pseudo-labeling issue in one-stage detection, where the model often fails to provide high-confidence pseudo labels in the initial stage. As shown in Fig.~\ref{conf_distri_fig}, we  analyze the pseudo-label distributions,   which show that the threshold of 0.4 can achieve a good trade-off between the quality and quantity of pseudo-labels.
 Meanwhile, other hyper-parameters like the weight of focal loss~\cite{lin2017focalloss} $\lambda$ are also set according to the training status of one-stage SSOD.  

Notably, these adaptions are shared  with   other SSOD methods for fair comparisons. Nevertheless, we still notice that it is  difficult to maximize the effectiveness of one-stage teacher-student learning based on  these adaptions alone.

\vspace{-1em}
\section{Experiments}
\subsection{Datasets and Metric}
We  validate the proposed OneTeacher  on two object detection datasets, namely COCO~\cite{coco} and Pascal VOC~\cite{voc}.  Specifically, COCO contains three splits, namely \textit{train2017}, \textit{val2017} and \textit{test2017}.  We build the label sets from \emph{train2017} with the percentages of    1\%, 2\%, 5\%, 10\% and 20\%  labeled data, respectively, and the  rest images  of \emph{train2017}  are used as the unlabeled sets. On all experiments  of COCO, we  evaluate the model  on the \textit{val2017}. For VOC, we use the VOC07 \textit{train}+\textit{val}  and the   VOC012 \textit{train}+\textit{val} as the labeled and unlabeled datasets, respectively. Based on  these settings, we further conduct  the experiments using additional  \textit{COCO20} as the unlabeled datasets.  The models are evaluated  on the VOC07 \textit{test}. For both COCO and VOC, we use \textit{AP{\tiny{50:95}}}, also known as \textit{mAP}, as the metric.

\subsection{Implementation Details} 
\label{detail}
\noindent\textbf{Fully-supervised baseline.} We  use YOLOv5-s~\cite{yolov5} as our base model,  and its  backbone is randomly initialized\footnote{Note that YOLOv5 is trained from scratch without  ImageNet~\cite{imagenet} pre-training.}. During  training,  the learning rate is set to 0.01 with a momentum of 0.937 and a weight decay of 0.0005.  The batch size is 64, and the total training steps are 500$k$,   of which 2$k$ steps  are for \textit{warm-up}. We also use \textit{EMA}~\cite{meanteacher-ssl-ts} to temporally ensemble the network parameters, and the  coefficient is set to 0.9999. We  keep all data augmentations in  YOLOv5, including random horizontal flip, mosaic~\cite{yolov4}, random image scale, random image translate and HSV color-space augmentation. 

\noindent\textbf{OneTeacher.} 
In {OneTeacher}, we remove the cosine learning rate decay, while  keeping  the  other    training configurations   the same as the fully-supervised baseline.    For semi-supervised learning,   we use SGD as the optimizer with a learning rate of  0.01, a momentum of 0.937 and a weight decay of 0.0005. The EMA coefficient is set to 0.9996. 
Training takes  500$k$ steps with 2$k$ warm-up steps,  and  the number of \textit{burn-up} steps~\cite{unbiasedteacher-ssod-ts} for the student network is  3$k$. The batch size for labeled data and unlabeled data are all set to 64.  As shown in Tab.~\ref{tab_adap}, we use random horizontal flip, mosaic, random image scale and random image translate as the weak data augmentation. The strong data augmentation for the student network includes  the weak ones  and color jittering, HSV color-space augmentation, grayscale, gaussian blur  and cutout. By default, the pseudo-labeling thresholds for regression and classification are set to 0.4 and 0.5, respectively. For MPR, the global threshold $\sigma_g$ is set to 0.25.   For the unsupervised loss, we employ the focal loss~\cite{lin2017focalloss} with $\alpha$ = 0.25 and $\gamma$ = 1.5.

\noindent\textbf{The compared methods.}  In addition to the supervised baseline, we also compare our method with the latest two-stage SSOD approach called \textit{Unbiased Teacher}~\cite{unbiasedteacher-ssod-ts} and a  representative    teacher-student  method  named by  \textit{STAC}~\cite{stac-ssod-ts}. We also  apply  the aforementioned adaptions  in Sec.~\ref{imp} to these baselines  to make them applicable  to YOLOv5.  The  other settings, such as learning rate, training steps and data augmentations, are   the same as {OneTeacher}. The confidence threshold $\sigma$ in the pseudo-labeling is set to 0.4 for both baselines. Since YOLOv5 uses EMA during training, we also keep this setting in the student network of STAC, denoted as \emph{\textbf{STAC+EMA}}.



\begin{table}[t]
	\centering
	\caption{\textbf{Comparisons of OneTeacher and baselines with ImageNet pre-training on COCO (1\%) and VOC.} For VOC, we use VOC07 \textit{train}+\textit{val} as the labeled data and VOC12 \textit{train}+\textit{val} as the unlabeled data.}
	\vspace{-1em}
	\begin{tabular}{lccccc}
		\toprule[1.2pt]
		\multicolumn{6}{c}{COCO (1\%)}                                                                                         \\ \midrule[0.8pt]
		\multicolumn{1}{l|}{Methods}          & mAP             & AP50             & APl             & APm             & APs            \\ \midrule[0.8pt]
		\multicolumn{1}{l|}{Supervised}       & 8.4\mypm{+0.0}  & 17.0\mypm{+0.0}  & 12.1\mypm{+0.0} & 9.1\mypm{+0.0}  & 3.0\mypm{+0.0} \\ \midrule[0.8pt]
		\multicolumn{1}{l|}{UbTeacher~\cite{unbiasedteacher-ssod-ts}} & 12.7\mypm{+4.3} & 23.7\mypm{+6.7}  & 16.9\mypm{+4.8} & 13.5\mypm{+4.4} & 4.8\mypm{+1.8} \\
		\multicolumn{1}{l|}{OneTeacher (ours)}       & \textbf{16.0}\mypm{+7.6}  & \textbf{28.2}\mypm{+11.2} &   \textbf{21.8}\mypm{+9.7}              &   \textbf{17.2}\mypm{+8.1}              &  \textbf{5.8}\mypm{+2.8}              \\ \bottomrule
		\toprule
		\multicolumn{6}{c}{VOC}                                                                                                         \\ \midrule[0.8pt]
		\multicolumn{1}{l|}{Methods}          & mAP             & AP50            & APl             & APm             & APs             \\ \midrule[0.8pt]
		\multicolumn{1}{l|}{Supervised}       & 43.5\mypm{+0.0}   & 71.2\mypm{+0.0} & 48.6\mypm{+0.0} & 31.0\mypm{+0.0} & 11.3\mypm{+0.0} \\ \midrule[0.8pt]
		\multicolumn{1}{l|}{UbTeacher~\cite{unbiasedteacher-ssod-ts}}        & 47.7\mypm{+4.2} & 75.1\mypm{+3.9} & 52.7\mypm{+4.1} & 33.7\mypm{+2.7} & \textbf{14.9}\mypm{+3.6} \\
		\multicolumn{1}{l|}{OneTeacher (ours)} & \textbf{50.0}\mypm{+6.5} & \textbf{76.1}\mypm{+4.9} & \textbf{55.8}\mypm{+7.2} & \textbf{36.2}\mypm{+5.2} & {14.4}\mypm{+3.1} \\ \bottomrule[1.2pt]
	\end{tabular}
	\label{IN}
\end{table}

\begin{table}[t]
	\centering
	\caption{\textbf{Performance comparison  of  YOLOv5-s, YOLOv5-m and  YOLOv5-x, which have different parameter scales. } OneTeacher is more  effective  for larger models.  }
	\vspace{-1em}
	\setlength\tabcolsep{6pt}
	\resizebox{0.5\textwidth}{!}{
		\begin{tabular}{ccccc}
			\toprule[1.2pt]
			\multicolumn{4}{c}{COCO (10\%)}                                                              \\ \midrule[0.8pt]
			\multicolumn{1}{l|}{Model}    & 		\multicolumn{1}{l|}{Params}                  & \multicolumn{1}{c|}{Methods}           & mAP  & AP50 \\ \midrule[0.8pt]
			\multicolumn{1}{c|}{\multirow{3}{*}{YOLOv5-s}} & 		\multicolumn{1}{c|}{\multirow{3}{*}{7.3M}} & \multicolumn{1}{c|}{Supervised}        & 21.8\mypm{+0.0} & 37.4\mypm{+0.0} \\
			\multicolumn{1}{c|}{}                          &		\multicolumn{1}{c|}{}    & \multicolumn{1}{c|}{Unbiased Teacher~\cite{unbiasedteacher-ssod-ts}}  & 25.3\mypm{+3.5} & 42.0\mypm{+2.6} \\
			\multicolumn{1}{c|}{}                       &  		\multicolumn{1}{c|}{}    & \multicolumn{1}{c|}{OneTeacher (ours)} & \textbf{29.1}\mypm{+8.2} & \textbf{45.3}\mypm{+7.9} \\
			\midrule[0.8pt]
			\multicolumn{1}{c|}{\multirow{3}{*}{YOLOv5-m}} &\multicolumn{1}{c|}{\multirow{3}{*}{21.4M}}& \multicolumn{1}{c|}{Supervised}        &  22.5\mypm{+0.0} & 37.9\mypm{+0.0} \\
			\multicolumn{1}{c|}{}                  &   		\multicolumn{1}{c|}{}        & \multicolumn{1}{c|}{Unbiased Teacher~\cite{unbiasedteacher-ssod-ts}}  & 28.4\mypm{+5.9} & 44.3 \mypm{+6.4} \\
			\multicolumn{1}{c|}{}      &        		\multicolumn{1}{c|}{}               & \multicolumn{1}{c|}{OneTeacher (ours)} & \textbf{33.3}\mypm{+10.8} & \textbf{49.7}\mypm{+11.8} \\ \midrule[0.8pt]
			\multicolumn{1}{c|}{\multirow{3}{*}{YOLOv5-x}} & \multicolumn{1}{c|}{\multirow{3}{*}{87.7M}}& \multicolumn{1}{c|}{Supervised}        & 23.7\mypm{+0.0} & 38.8\mypm{+0.0} \\
			\multicolumn{1}{c|}{}          &         		\multicolumn{1}{c|}{}          & \multicolumn{1}{c|}{Unbiased Teacher~\cite{unbiasedteacher-ssod-ts}}  & 36.5\mypm{+14.8} & 52.1\mypm{+16.3} \\
			\multicolumn{1}{c|}{}               &   		\multicolumn{1}{c|}{}           & \multicolumn{1}{c|}{OneTeacher (ours)} & \textbf{38.9}\mypm{+17.2} & \textbf{55.3}\mypm{+19.5} \\ \bottomrule[1.2pt]
	\end{tabular}}
	\label{yolox}
\end{table}

\subsection{Experimental Results}
\subsubsection{Comparison with existing methods}
We first compare {OneTeacher} with the   supervised baseline and existing SSOD methods  on  COCO~\cite{coco} and  VOC~\cite{voc},  of which results are given in   Tab.~\ref{sota} - \ref{yolox}. 

\label{sota_dis}
Tab.~\ref{sota} shows the performance comparison on COCO at different label scales.  We can observe that  {OneTeacher} achieves distinct performance gains over  the  supervised baseline, \emph{e.g.,} +33.5\%  (relative improvement) on 10\% COCO labeled data.  Such an improvement is  even more significant than that of  two-stage   SSOD~\cite{softteacher-ssod-ts,unbiasedteacher-ssod-ts,stac-ssod-ts}, \emph{e.g.}, +32\% by Unbiased Teacher on FasterRCNN~\cite{fasterrcnn}. 
Compared to existing SSOD methods, the advantage of {OneTeacher} is also very obvious, \emph{e.g.}, up to +15\% than Unbiased Teacher (UbTeacher) and up to +59\% than STAC.  Meanwhile, we also notice that these SSOD baselines  do not perform as well on YOLOv5 as they do on FasterRCNN~\cite{unbiasedteacher-ssod-ts,stac-ssod-ts}. STAC, in particular, performs even worse than the supervised  scheme  when deploying its   default two-stage setup. These results show that YOLOv5 is a strong and challenging  base model, whose training techniques partially offset the benefit of SSOD.  Meanwhile, we also notice that when using  the  less COCO labeled data, the benefits of OneTeacher and other SSOD  methods   become relatively smaller, as shown in Tab.~\ref{sota}.  We attribute this problem to the random  parameter initialization of YOLOv5~\cite{yolov5}. Without ImageNet pre-training, YOLOv5 is prone to over-fitting  to a small amount of  labeled data. In this case,     we also provide the   experimental results with ImageNet pre-training in Tab.~\ref{IN},  which can prove  the effectiveness of {OneTeacher} on  smaller datasets.   From Tab.~\ref{IN}, we can see that OneTeacher achieves significant performance gains over  baselines  on less label data, \emph{e.g.,} +3.3 mAP than UbTeacher on COCO (1\%). On VOC, the merits of OneTeacher are also obvious, \emph{e.g.,} +2.3 mAP than UbTeacher. These results well  confirm the effectiveness of OneTeacher with ImageNet pre-training. Although OneTeacher works well with ImageNet pre-training, we still report the results with  the default random initialization setting  in our paper, which  is more challenging.

\begin{figure*}[t]
	\centering
	\includegraphics[width=\textwidth]{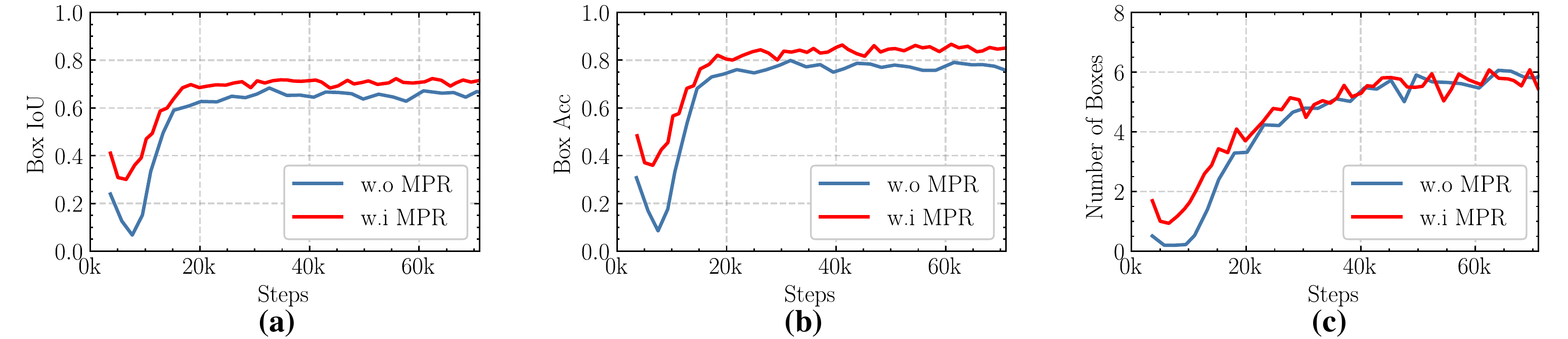}
	\vspace{-2em}
	\caption{\textbf{Comparison of OneTeacher with and without MPR in terms of the averaged IoU score (a), accuracy (b) and number (c) of pseudo-labels  on COCO (10\%).} MPR can significantly improve the quality of pseudo-labels.  }
	\label{MPR_fig}
\end{figure*}

\begin{table}[t]
	\centering
	\caption{\textbf{Ablation Study  of Multi-view Pseudo-Label Refinement (MPR) and Decoupled Semi-supervised Optimization (DSO) on {COCO (10\%)}. }  }
	\footnotesize 
	{
		\setlength
		\tabcolsep{22pt}
		\begin{tabular}{cccc}
			\toprule[1.2pt]
			\multicolumn{4}{c}{COCO (10\%)}                          \\ \midrule[0.8pt]
			\multicolumn{1}{c|}{MPR} & \multicolumn{1}{c|}{DSO} & mAP  & AP50 \\ \midrule[0.8pt]
			\multicolumn{1}{c|}{$\times$}    & \multicolumn{1}{c|}{$\times$}    & 25.3 & 42.0 \\
			\multicolumn{1}{c|}{$\checkmark$}    & \multicolumn{1}{c|}{$\times$}    & 26.1 & 43.1 \\
			\multicolumn{1}{c|}{$\times$}    & \multicolumn{1}{c|}{$\checkmark$}    & 28.1 & 44.0 \\ \midrule[0.8pt]
			\multicolumn{1}{c|}{$\checkmark$}    & \multicolumn{1}{c|}{$\checkmark$}    & 29.1 & 45.3 \\ \bottomrule[1.2pt]
	\end{tabular} }
	\label{ablation}
\end{table}

\begin{table}[t]
	\centering
	\caption{\textbf{Effects of different pseudo-labeling  thresholds in decoupled semi-supervised optimization.}  $t_r$ and $t_c$  are thresholds for regression and classification, respectively. }
	\footnotesize 
	{
		\setlength
		\tabcolsep{10pt}
		\begin{tabular}{ccccccc}
			\toprule[1.2pt]
			\multicolumn{7}{c}{COCO (10\%)}                                               \\ \midrule[0.8pt]
			\multicolumn{1}{c|}{$t_r$}  & \multicolumn{1}{c|}{$t_c$}  & mAP  & AP50 & APl  & APm  & APs  \\ \midrule[0.8pt]
			\multicolumn{1}{c|}{0.2} & \multicolumn{1}{c|}{0.3} & 27.7 & 43.7 & 36.3 & 30.2 & 14.3 \\
			\multicolumn{1}{c|}{0.2} & \multicolumn{1}{c|}{0.4} & 27.8 & 43.5 & 36.6 & 30.6 & 14.7 \\
			\multicolumn{1}{c|}{0.2} & \multicolumn{1}{c|}{0.5} & 28.3 & 44.4 & 36.7 & 31.7 & 14.2 \\
			\multicolumn{1}{c|}{0.3} & \multicolumn{1}{c|}{0.3} & 27.9 & 43.7 & 36.8 & 30.3 & 14.8 \\
			\multicolumn{1}{c|}{0.3} & \multicolumn{1}{c|}{0.4} & 28.8 & 45.0 & 37.6 & 31.7 & 15.1 \\
			\multicolumn{1}{c|}{0.3} & \multicolumn{1}{c|}{0.5} & 29.0 & 45.4 & 37.6 & 32.4 & 14.7 \\
			\multicolumn{1}{c|}{0.4} & \multicolumn{1}{c|}{0.4} & 28.1 & 43.9 & 37.3 & 30.5 & 14.3 \\
			\multicolumn{1}{c|}{\textbf{0.4}} & \multicolumn{1}{c|}{\textbf{0.5}} & \textbf{29.1} & \textbf{45.3} & \textbf{38.8} & \textbf{31.8} & \textbf{14.6 }\\
			\multicolumn{1}{c|}{0.5} & \multicolumn{1}{c|}{0.3} & 28.7 & 44.9 & 38.0 & 31.5 & 14.8 \\
			\multicolumn{1}{c|}{0.5} & \multicolumn{1}{c|}{0.4} & 28.6 & 44.5 & 37.7 & 31.3 & 15.0 \\ \bottomrule[1.2pt]
	\end{tabular}}
	\label{thre}
\end{table}

\begin{table}[t]
	\centering
	\caption{\textbf{Comparisons of our design with other candidates.  AR denotes  the augmented-view refinement in  MPR.} The first block compares the default augmentation strategy of UbTeacher with our adaption. The second block compares the common flip ensemble~\cite{tang2021humbleteacher-ssod-ts}   with AR. }
	\setlength\tabcolsep{20pt} 
	{\begin{tabular}{l|cc}
			\toprule[1.2pt]
			\multicolumn{3}{c}{COCO (10\%)}                          \\ \midrule[0.8pt]
			\multirow{1}{*}{Settings}         & \multirow{1}{*}{mAP} & \multirow{1}{*}{AP50} \\  \midrule[0.8pt] 
			+Default Aug~\cite{yolov5}                      &     12.6                 &       24.1                \\
			\multicolumn{1}{l|}{+Adapted Aug (Ours)}  &     26.1                 & \multicolumn{1}{c}{43.1}  \\ \midrule[0.8pt]
			+Flip Ensemble~\cite{tang2021humbleteacher-ssod-ts}                                 &  24.8                    &   41.4     \\
			+AR (ours)                                            &  26.1                    &     43.1           \\ \bottomrule[1.2pt]
	\end{tabular}}
	\label{ablationv2}
\end{table}

  The results of VOC are given in Tab.~\ref{voc}.  Compared to COCO, VOC  has fewer object categories and  the visual scenes involved are relatively simpler. In this case, the fully supervised baseline can already achieve good performance with limited label information, \emph{i.e.,} 41.3 mAP.   We also observe that the performance gains of UbTeacher are not significant on VOC07+12, \emph{i.e.,} +0.8 mAP against the supervised baseline.  We conjectured that the reasons are two-fold. Firstly, the fully supervised baseline is  highly competitive due to its advanced training techniques.  Secondly, these SSOD baselines still suffer the problems of pseudo-label noise and optimization conflict.   Compared to these SSOD baselines, OneTeacher  still demonstrates superior performance than the supervised baseline, \emph{i.e.}, +4.8 mAP.

   We  further apply {OneTeacher}  to  a larger one-stage network, \emph{i.e.,} YOLOv5-x,  of which results are  reported in Tab.~\ref{yolox}.  A first observation   is that  YOLOv5-x  is prone to over-fitting on a small label set, resulting in worse performance than  YOLOv5-s. However, when {OneTeacher} is  applied, the performance of YOLOv5-x  boosts  from 21.7 mAP to 38.9 mAP, which is much more significant than that of YOLOv5-s.  Additionally, {OneTeacher} also outperforms {Unbiased Teacher}~\cite{unbiasedteacher-ssod-ts} on YOLOv5-x. These results  greatly validate the     generalization ability of {OneTeacher} on large detection  networks.

  Overall, these results not only show the challenges of YOLOv5, but also well confirm the effectiveness of OneTeacher towards one-stage SSOD.

\subsubsection{Ablation study.}
We further ablate two key designs of OneTeacher, \emph{i.e.}, \textit{Multi-view Pseudo-label Refinement} (MPR) and \textit{Decoupled Semi-supervised Optimization}  (DSO) on  COCO,  of which results are given  in Tab.~\ref{ablation}.  We can find that all  designs of  these two schemes are beneficial to model performance. Among them, the benefit of   DSO is the most significant, which subsequently confirms the issue of multi-task optimization conflict   we argued. Meanwhile, when combining  MPR and DSO, the performance can be improved by up to 15\%,  suggesting the validity of these two  schemes.

\begin{figure}[t]
	\centering
	\includegraphics[width=0.35\textwidth]{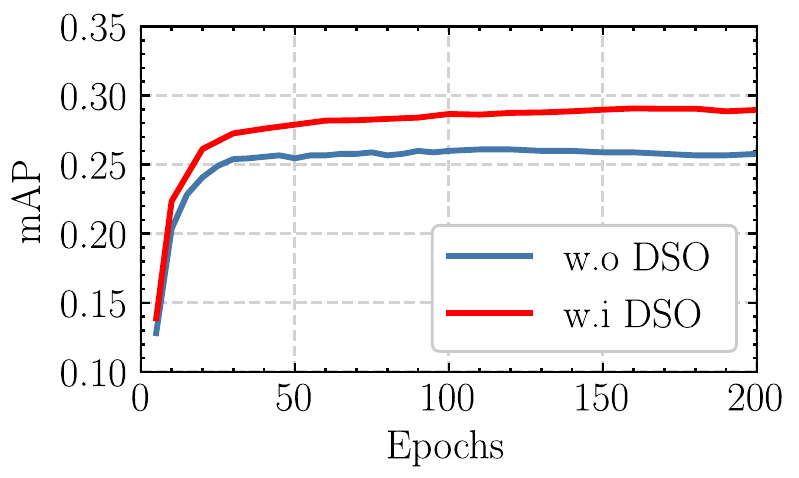} 
	\vspace{-1em}
	\caption{\textbf{The training process of OneTeacher with and without DSO.} DSO can address the conflicts of the coupled multi-task optimization and significantly improve the efficiency of semi-supervised learning.  }
	\label{DSO_fig}
	\vspace{-1em}
\end{figure}

To better understand the impact of MPR and DSO  on semi-supervised learning, we further  visualize their training processes  in  Fig.~\ref{MPR_fig} and  Fig.~\ref{DSO_fig}.   In Fig.~\ref{MPR_fig}, we compare the quality of pseudo-boxes of {OneTeacher}  with and without MPR. The first observation is that with the help of MPR, the quality of pseudo-boxes is significantly improved, especially in the initial  training phase. For instance, at about 4$k$ steps, the averaged IoU and accuracy scores are almost doubled. Meanwhile, the number of high-quality pseudo-boxes is also  obviously  increased. These results strongly confirm the effectiveness of MPR in tackling with the low-quality pseudo-label problem  of  one-stage SSOD.  Fig.~\ref{DSO_fig} shows that DSO can greatly improve the training efficiency of OneTeacher,  which also  helps the base model achieve better performance.

 In Tab.~\ref{thre}, we also validate the effects of different pseudo-labeling thresholds in DSO.    The first observation from this table is that the optimal thresholds for YOLOv5 are much smaller than those for Faster-RCNN, \emph{e.g.}, 0.4 \emph{v.s.} 0.7, reflecting the difference between one-stage and two-stage SSODs. More importantly, we can also observe that setting different pseudo-labeling thresholds for regression and classification is indeed beneficial for one-stage SSOD, \emph{e.g.}, +0.9mAP, suggesting that these two tasks in YOLOv5  have different tolerances about pseudo-label noises.

\begin{figure*}[!t]
	\centering
	\includegraphics[width=1\textwidth]{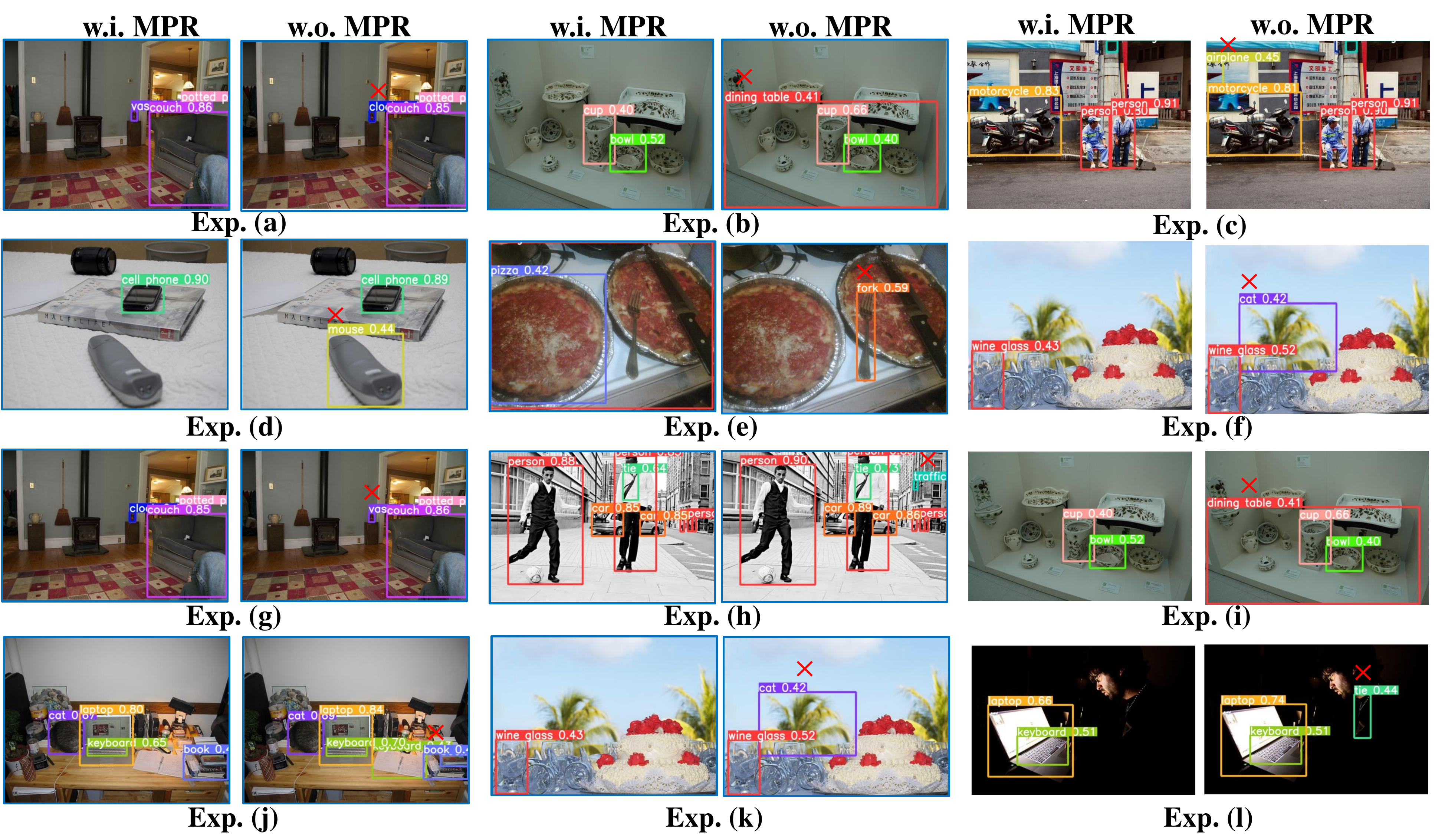}
	\vspace{-2em}
	\caption{\textbf{Visualizations of the   pseudo-labels generated by OneTeacher with and without MPR.}  With the help of MPR, OneTeacher can obtain more  high-quality   pseudo-labels.  }
	\label{vis}
\end{figure*}
 Tab.~\ref{ablationv2} reports the results of the alternative designs for OneTeacher.  In  first block, we compare   two types of data augmentation settings for YOLOv5.  We observe that compared to the  default augmentation scheme used in two-stage SSODs like UbTeacher~\cite{unbiasedteacher-ssod-ts}, our new deployment for YOLOv5  greatly improves the model performance, \emph{i.e.,} +13.5 mAP.  This result  indicates that the original augmentations of YOLOv5  greatly contribute to its  performance.  In this case,  the   weak augmentation setting   of UbTeacher, which does not include YOLOv5's augmentation  methods, will significantly degenerate SSOD performance  of YOLOv5.   In Tab.~\ref{ablationv2}, we also compare the augmented-view refinement (AR) in MPR with its alternative called \textit{Flip Ensemble}~\cite{tang2021humbleteacher-ssod-ts}, which refines pseudo-labels via directly averaging  the  predictions of the original view and flipped view.  It can be seen that  the  benefit  of Flip Ensemble is much less significant than that of our augmented-view refinement.  One possible reason is that this approach will  ensemble all views of predictions without comparisons and filtering, thereby introducing more label noises to one-stage SSOD.   Instead, the proposed AR only fuses the certain predictions from two different views, which can effectively improve the quality of pseudo-labels.

\subsubsection{Qualitative Analysis}
We visualize the  pseudo-labels  produced  by  \textit{OneTeacher} with and without MPR  in Fig.~\ref{vis}. From these examples, we first observe that MPR can effectively filter  the incorrect detections, \emph{e.g.}, the  ``\textit{dining table}'' in  Exp. (b).  Meanwhile, the pseudo-boxes with incorrect category predictions can also be refined by MPR, \emph{e.g.},  the wrong prediction of  ``\textit{clock}'' in  Exp. (a)  is corrected to  ``\textit{vase}''  by MPR. Besides, MPR can also bring more high-quality pseudo-labels to OneTeacher, \emph{e.g.}, the missed detection ``\textit{pizza}''  in  Exp. (e) is also detected after the refinement of  MPR. These results well confirm  the effectiveness of MPR towards the issue of low-quality pseudo-labeling.  

\section{Conclusion}
In this paper, we  focus on one-stage semi-supervised object detection (SSOD), which is often overlooked in existing literature. Specifically, we  identify its two key challenges compared to two-stage SSOD,  namely   \textit{low-quality pseudo-labeling} and \textit{multi-task optimization conflict}. To address these issues, we   present   a novel  teacher-student learning  paradigm  for one-stage  detection networks, termed \textit{OneTeacher}. 
To  handle the first issue, OneTeacher  applies a novel  design called  \emph{Multi-view Pseudo-label Refinement} (MPR) to improve pseudo-labels quality from micro- to macro-views. Meanwhile,  OneTeacher  adopts a \emph{Decoupled Semi-supervised Optimization} (DSO)  scheme to address the  multi-task optimization conflicts via a branching structure  and task-specific pseudo-labeling.   In addition, we also apply the advanced one-stage detection network YOLOv5 as our base model and carefully revise its implementation to maximize the benefits of SSOD. To validate OneTeacher, we   conduct extensive experiments on {COCO} and {Pascal VOC}. The experimental results show that OneTeacher greatly outperforms both supervised and  semi-supervised  methods under  different settings, and  also confirm its effectiveness  towards the  aforementioned  key issues of one-stage SSOD. 

\section*{Acknowledgment}
This work was supported by the National Science Fund for Distinguished Young Scholars (No.62025603), the National Natural Science Foundation of China (No. U21B2037, No. 62176222, No. 62176223, No. 62176226, No. 62072386, No. 62072387, No. 62072389, and No. 62002305), Guangdong Basic and Applied Basic Research Foundation (No.2019B1515120049), and the Natural Science Foundation of Fujian Province of China (No.2021J01002).   We also thank
Huawei Ascend Enabling Laboratory for the continuous support
in this work. Particularly, we highly appreciate Qing Xu and Rongrong Fu for their efforts and valuable suggestions to this paper. 



\bibliographystyle{IEEEtran}
\bibliography{IEEEabrv,mybstfile}

\begin{IEEEbiography}[{\includegraphics[width=1in,height=1.25in,clip,keepaspectratio]{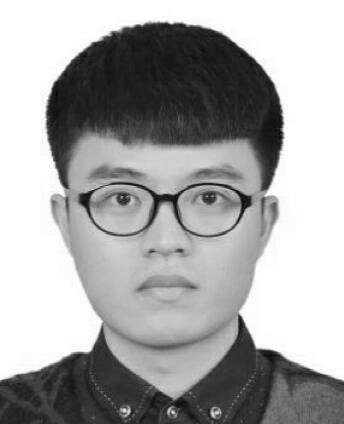}}]{Gen Luo}  is currently pursuing the phd’s
	degree in Xiamen University.
	His research interests include vision-and-language learning and semi-supervised learning.
\end{IEEEbiography}

\begin{IEEEbiography}[{\includegraphics[width=1in,height=1.25in,clip,keepaspectratio]{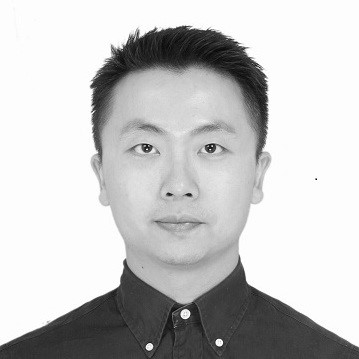}}]{Yiyi Zhou}  received the Ph.D. degree from Xiamen University, China, in 2019, under the supervision of Prof. Rongrong Ji. He was a Postdoctoral Research Search Fellow with Xiamen University from 2019 to 2022. He is currently an associate professor at School of Informatics and Institute of Artificial Intelligence of Xiamen University. His research interests include multimedia analysis and computer vision.
\end{IEEEbiography}

\begin{IEEEbiography}[{\includegraphics[width=1in,height=1.25in,clip,keepaspectratio]{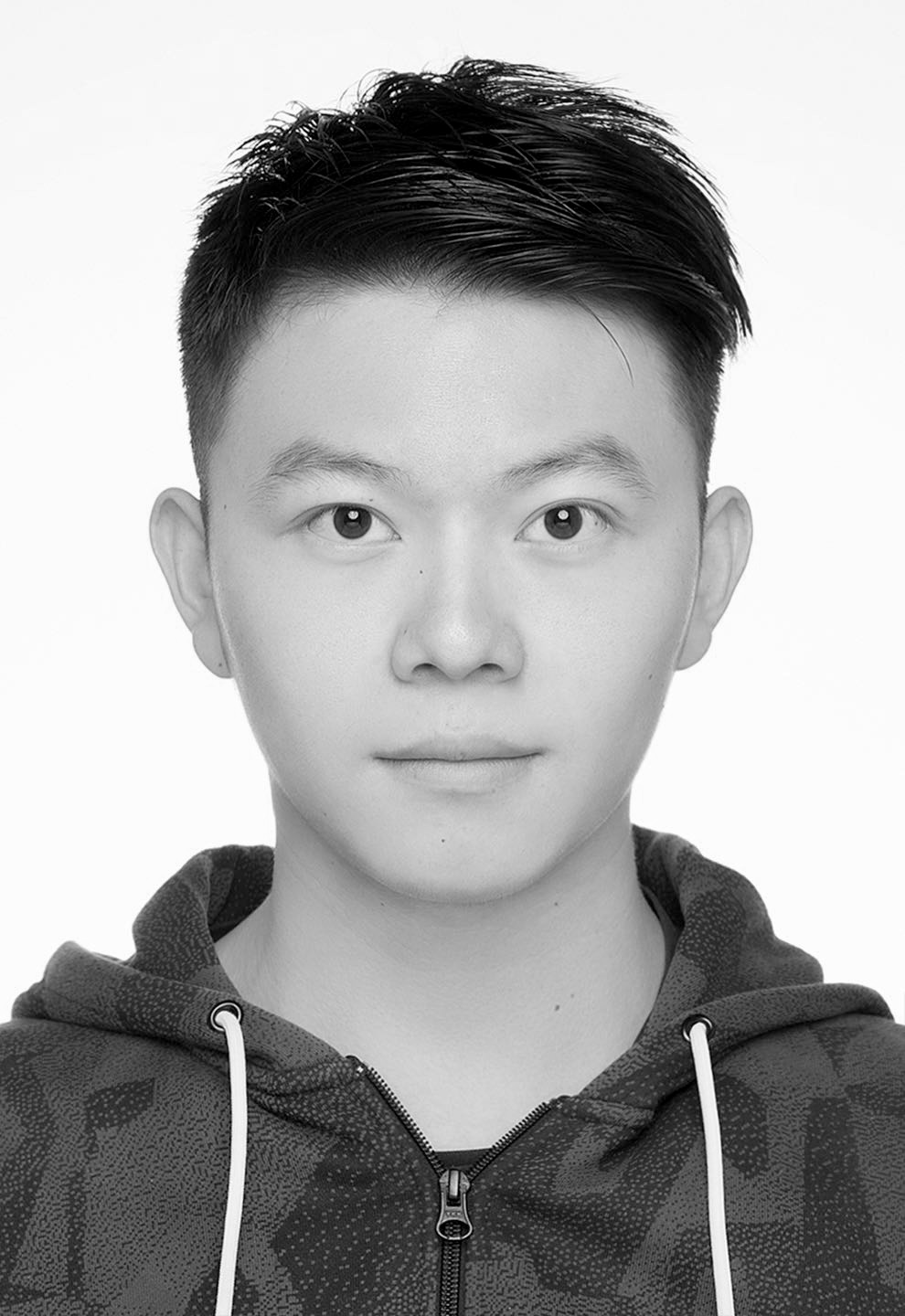}}]{Lei Jin} is a postgraduate of the School of Informatics and a member of Media Analytics and Computing (MAC) lab  of Xiamen University, China.
\end{IEEEbiography}

\begin{IEEEbiography}[{\includegraphics[width=1in,height=1.25in,clip,keepaspectratio]{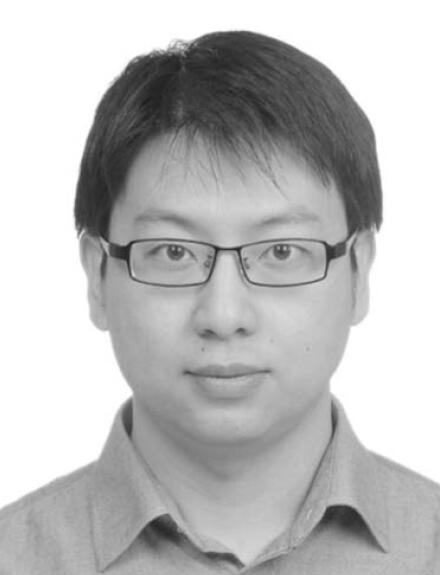}}]{Xiaoshuai Sun} (Senior Member, IEEE)
	received the B.S. degree in computer
	science from Harbin Engineering University, Harbin,
	China, in 2007, and the M.S. and Ph.D. degrees in
	computer science and technology from the Harbin
	Institute of Technology, Harbin, in 2009 and 2015,
	respectively. He was a Postdoctoral Research Fellow
	with the University of Queensland from 2015 to
	2016. He served as a Lecturer with the Harbin
	Institute of Technology from 2016 to 2018. He is
	currently an Associate Professor with Xiamen University,
	China. He was a recipient of the Microsoft Research Asia Fellowship in 2011.
\end{IEEEbiography}

\begin{IEEEbiography}[{\includegraphics[width=1.1in,height=1.25in,clip,keepaspectratio]{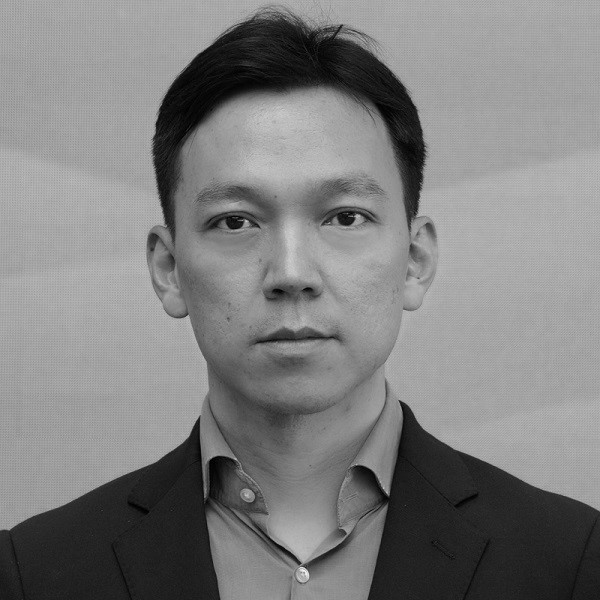}}]{Rongrong Ji}
	(Senior Member, IEEE)   is a Nanqiang Distinguished Professor at Xiamen University, the Deputy Director of the Office of Science and Technology at Xiamen University, and the Director of Media Analytics and Computing Lab. He was awarded as the National Science Foundation for Excellent Young Scholars (2014), the National Ten Thousand Plan for Young Top Talents (2017), and the National Science Fundation for Distinguished Young Scholars (2020). His research falls in the field of computer vision, multimedia analysis, and machine learning. He has published 50+ papers in ACM/IEEE Transactions, including TPAMI and IJCV, and 100+ full papers on top-tier conferences, such as CVPR and NeurIPS. His publications have got over 10K citations in Google Scholar. He was the recipient of the Best Paper Award of ACM Multimedia 2011. He has served as Area Chairs in top-tier conferences such as CVPR and ACM Multimedia. He is also an Advisory Member for Artificial Intelligence Construction in the Electronic Information Education Commitee of the National Ministry of Education.
\end{IEEEbiography}
\end{document}